\pdfoutput=1

\documentclass[11pt]{article}
\usepackage[]{ACL2023}

\usepackage{times}
\usepackage{latexsym}
\usepackage[T1]{fontenc}
\usepackage[utf8]{inputenc}
\usepackage{microtype}
\usepackage{inconsolata}
\usepackage{graphicx}
\usepackage{booktabs} 
\usepackage{multirow} 
\usepackage{amsmath} 
\usepackage{amssymb}
\usepackage{tikz}
\usepackage{pgfplots}
\usepackage{pgf-pie}
\usepackage{amsfonts}
\usepackage{pifont}
\definecolor{category1}{RGB}{255,153,136}
\definecolor{category2}{RGB}{247,196,127}
\definecolor{category3}{RGB}{255,227,148}
\pgfplotsset{compat=1.18} 
\usepackage[utf8]{inputenc} 
\usepackage[T1]{fontenc}    
\usepackage{hyperref}       
\usepackage{url}            
\usepackage{booktabs}       
\usepackage{amsfonts}       
\usepackage{nicefrac}       
\usepackage{microtype}      
\usepackage{xcolor}         
\usepackage{graphicx} 
\usepackage{amsmath} 
\usepackage{multirow}
\usepackage{subcaption}

\title{NOTA: Multimodal Music Notation Understanding for  Visual Large Language Model}
\author{Mingni Tang$^{1,\dag}$, Jiajia Li$^{2,3,\dag}$, Lu Yang$^{1}$, Zhiqiang Zhang$^{1}$, Jinghao Tian$^{1}$,\\ \textbf{Zuchao Li$^{1,}$\thanks{$\ $  Corresponding author. This work was supported by the National Natural Science Foundation of China (No. 62306216, No. 72074171, No. 72374161), the Technology Innovation Program of Hubei Province (Grant No. 2024BAB043), the Natural Science Foundation of Hubei Province of China (No. 2023AFB816).}, 
  Lefei Zhang$^{1}$, Ping Wang$^{2,3,*}$}\\
$^{1}$School of Computer Science, Wuhan University, Wuhan, China\\
$^{2}$Key Laboratory of Archival Intelligent Development and Service, NAAC \\
$^{3}$School of Information Management, Wuhan University, Wuhan, China \\
{\tt \{minnie-tang, cantata, yang\_lu, zhangzhiqiang, jinhaotian,}\\{ \tt zcli-charlie, zhanglefei, wangping\}@whu.edu.cn}
}

\begin{document}
\maketitle
\begin{abstract}
Symbolic music is represented in two distinct forms: two-dimensional, visually intuitive score images, and one-dimensional, standardized text annotation sequences.
While large language models have shown extraordinary potential in music, current research has primarily focused on unimodal symbol sequence text.
Existing general-domain visual language models still lack the ability of music notation understanding. Recognizing this gap, we propose NOTA, the first large-scale comprehensive multimodal music notation dataset. It consists of 1,019,237 records, from 3 regions of the world, and contains 3 tasks. Based on the dataset, we trained NotaGPT, a music notation visual large language model. Specifically, we involve a pre-alignment training phase for cross-modal alignment between the musical notes depicted in music score images and their textual representation in ABC notation. Subsequent training phases focus on foundational music information extraction, followed by training on music notation analysis. Experimental results demonstrate that our NotaGPT-7B achieves significant improvement on music understanding, showcasing the effectiveness of NOTA and the training pipeline. Our datasets are open-sourced at \url{https://huggingface.co/datasets/MYTH-Lab/NOTA-dataset}.
\end{abstract}
\section{Introduction} \label{introduction}
Music is expressed primarily in two forms: auditory music and symbolic music. Symbolic music can be represented in two-dimensional space through scores that display notes, rhythms, and dynamics, thereby guiding performers on how to play the music. It can also be expressed through lines of text sequences, effectively linearizing the complexity of music for ease of computer processing and programmatic manipulation. The evolution of Natural Language Processing (NLP) and multimodal interactions has provided valuable insights into the understanding and generation of music.
With the advent of universal dialogue Multimodal Large Language Models (MLLMs) such as GPT-4\citep{gpt4}, specialized models designed for various professional domains~\citep{dey-etal-2024-socialite,baez-saggion-2023-lsllama}, including music (e.g., MU-LLaMA~\citep{liu2024music}), have begun to proliferate. 
However, these works have only focused on the single modality of text, and in order to interact with multiple modalities, some MLLMs have been recently introduced. 
Nevertheless, these MLLM models mainly focus on the task of multimodal information extraction in the general domain, and rarely involve multimodal information extraction.
Most existing datasets focus on specific symbols or audio (like ABC notation~\citep{Nottingham}, MIDI~\citep{ryu2024mid}, WAV~\citep{sturm2013gtzan}, and lyrics~\citep{ccano2017moodylyrics}) and do not emphasize the visual modality, limiting their ability to enable MLLMs to understand music notation.
Visual representations serve as a tangible record of music. These images not only encapsulate the score's information but also visually delineate its intricate structures~\citep{tjh1,tjh2}.  

\begin{figure*}[ht]
\centering
\small

\begin{minipage}{0.3\textwidth}
  \includegraphics[width=\textwidth]{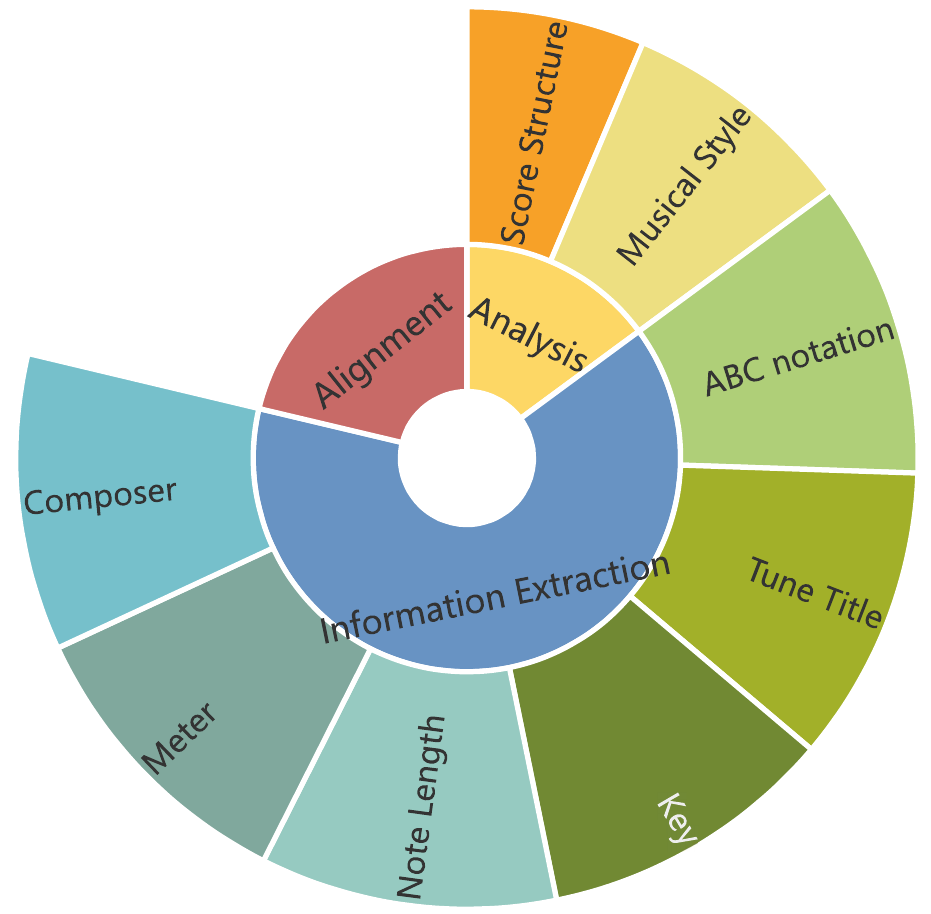}
\end{minipage}
\hspace{0.05\textwidth} 
\begin{minipage}{0.29\textwidth}
\resizebox{\textwidth}{!}{
\begin{tabular}{ccc}
\toprule
\multicolumn{2}{c}{Train Dataset} & \\ 
       \midrule
        Cross-modal Alignment & 28,125 & \\
        \midrule
         \multicolumn{2}{c}{\textit{Music Information Extraction}} \\ 
        T (Tune Title) & 161,633 & \\
        K (Key) & 161,633 & \\
        L (Unit Note Length) & 161,633 & \\
        M (Meter) & 161,633 & \\
        C (Composer) & 161,633 & \\
        ABC notation & 161,636 & \\
        \midrule
        \multicolumn{2}{c}{\textit{Music Notation Analysis}} \\ 
        Score Structure & 150 & \\
        Musical Style & 300 & \\
 \bottomrule
\end{tabular} 
}
\end{minipage}
\hspace{0.05\textwidth} 
\begin{minipage}{0.27\textwidth}
\resizebox{\textwidth}{!}{
\begin{tabular}{ccc}
\toprule
\multicolumn{2}{c}{Test Dataset} & \\ 
       \midrule
        Region Bias Test & 9,150 & \\
        \midrule
       \multicolumn{2}{c}{\textit{Music Information Extraction}} \\ 
        T (Tune Title) & 1,851 & \\
        K (Key) & 1,851 & \\
        L (Unit Note Length) & 1,851 & \\
        M (Meter) & 1,851 & \\
        C (Composer) & 1,851 & \\
        ABC notation & 1,851 & \\
        \midrule
        \multicolumn{2}{c}{\textit{Music Notation Analysis}} \\ 
        Score Structure & 300 & \\
        Musical Style & 400 & \\
 \bottomrule
\end{tabular} 
}
\end{minipage}
    \caption{Data distribution of NOTA dataset.}
    \label{dataset construction}
\end{figure*}

To address the above limitations, we introduce NOTA, the first and largest comprehensive dataset designed to train and evaluate multimodal models in music notation understanding. Spanning three distinct global regions, NOTA encompasses over 1 million records of music scores. And it is structured around 3 pivotal tasks: music information extraction, cross-modal alignment test, and music notation analysis. These tasks cover various aspects of music, including music theory, composition, genres, musical ontological elements, and humanistic connotations. Our dataset is divided into two main parts: the training dataset and the test dataset. On the one hand, it provides training materials for researchers in the community to train their own multimodal music models. On the other hand, it enables the evaluation of existing multimodal models' ability to understand music.

Based on this dataset, we trained a 7B model, NotaGPT, capable of understanding music notation across multiple modalities, including visual modalities.
This training process comprises a pre-alignment training focused on cross-modal alignment between the visual symbols in the music scores and their textual symbolic counterparts. This is followed by fine-tuning that aim at foundational music information extraction, and music notation analysis. 

Utilizing NOTA, we conducted comprehensive experiments on 17 mainstream multimodal large language models. Specifically, we input music score images and background information about the pieces, asking them to output basic information such as note lengths and key signatures or to perform analyses of the musical style and rhythm. Even the best-performing model, Gemini, achieved a music information extraction rate of only 33.34\%. In contrast, our 7B model, trained on our dataset, achieved 67.84\%. The experimental results demonstrate the limitations in model performance caused by the lack of multimodal music datasets and highlight the effectiveness of our NOTA dataset and our training pipeline.

Our contribution can be summarized as follows: We introduced NOTA, the first and largest comprehensive multimodal music notation understanding dataset. This dataset encompasses 1,019,237 records from 3 distinct global regions and is dedicated to 3 tasks, addressing the resource limitation available for multimodal music notation understanding.

\section{Related Work}

\subsection{Multimodal Benchmark}

In the fields of NLP and multimodal interactions, traditional evaluation metrics predominantly focus on assessing specific capabilities of a model within singular task types\citep{goyal2017making}. For example, the GLUE (General Language Understanding Evaluation)~\citep{glue} benchmark is a collection of diverse natural language understanding tasks designed to evaluate and advance the performance of models on a wide range of language comprehension challenges.
These criteria either provide more dimensions of assessment ~\citep{guha2024legalbench,sun2024scieval}and advanced capabilities or employ sophisticated evaluation mechanisms~\citep{wang2023pandalm,valmeekam2024planbench}. 
For instance, the C-Eval~\citep{huang2024c} benchmark addresses the gap in Chinese language data.

The evolution of evaluation benchmarks in NLP and multimodal fields has consequently influenced the benchmarks used in music evaluation. Presently, music evaluation metrics generally concentrate on distinct musical capabilities, such as music generation~\citep{agostinelli2023musiclm,melechovsky2023mustango}, music information retrieval~\citep{kong2020panns,zhao2021musicoder} and music understanding~\citep{ZIQI-Eval}. 
Some initiatives, such as ChatMusician~\citep{yuan2024chatmusician}, attempt to unify tasks in music generation and comprehension, yet suffer from limited data volumes.
Despite the rapid development of multimodal generative models, there is still a lack of data and benchmarks that can effectively evaluate the models' capabilities in understanding visual modality of music score images.

\begin{figure*}
    \centering
    \hspace*{-0.18cm} %
    \includegraphics[width=1\linewidth]{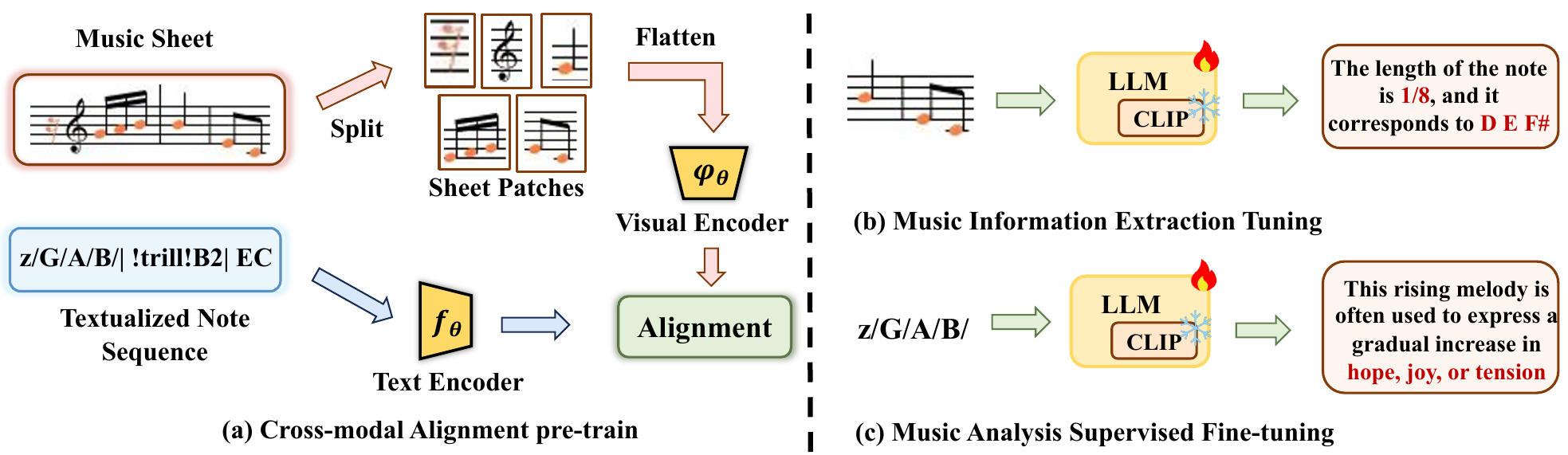}
    \caption{The figure shows the three-phase training process for NotaGPT-7B.}
    \label{fig:framework}
\end{figure*}
\subsection{Generative Models for Music Understanding and Generation}

With the advent of generative dialogue LLMs such as ChatGPT\citep{chatgpt}, alongside a series of universal dialogue MLLMs, specialized models designed for various professional domains~\citep{li2024llava,sun2022ringmo}, including music (e.g., MU-LLaMA~\citep{liu2024music}), have begun to proliferate.
As these MLLMs continue to evolve, music understanding capabilities have also been enhanced. For instance, current models like MusicAgent~\citep{yu-etal-2023-musicagent} and MusicLM~\citep{agostinelli2023musiclm} have made remarkable progress in music comprehension and generation abilities.

Generative models for music understanding and generation can be broadly categorized into two modalities: audio music~\citep{huangmulan,copet2024simple} and symbolic music~\cite{tian2024n,lu2023musecoco}. The former predominantly incorporates audio modalities into large language models~\citep{huang2024audiogpt} or employs diffusion models (e.g., JEN-1~\citep{li2023jen} and MeLoDy~\citep{lam2024efficient}) to process the audio components of music; the latter typically converts symbolic music information into sequences for integration into large language models~\citep{yuan2024chatmusician,geerlings2020interacting}.

The efficacy of these models hinges on precise instruction fine-tuning and cross-modal alignment~\cite{geerlings2020interacting}, utilizing specific musical datasets. Nevertheless, current generative music LLMs lack the ability to understand images of music scores in the visual modality.

\subsection{Multimodal information extraction}
Multimodal information extraction first searches for alignment in the two modalities connects them together, and then performs information extraction. It can be divided into two main categories: visual entity extraction and visual event extraction. In MORE~\citep{more}, the objective is to predict relations between objects and entities based on both textual and image inputs. Visual event extraction can be further divided into situation recognition~\citep{situation} and grounded situation recognition~\citep{grounded}.
With the development of MLLMs, information extraction datasets for different tasks have also evolved~\citep{wan2021fl,yuan2023joint}. However, there is still a lack of multimodal information extraction models and datasets specifically for the music domain.
\begin{figure*}
    \centering
    \includegraphics[width=1\linewidth]{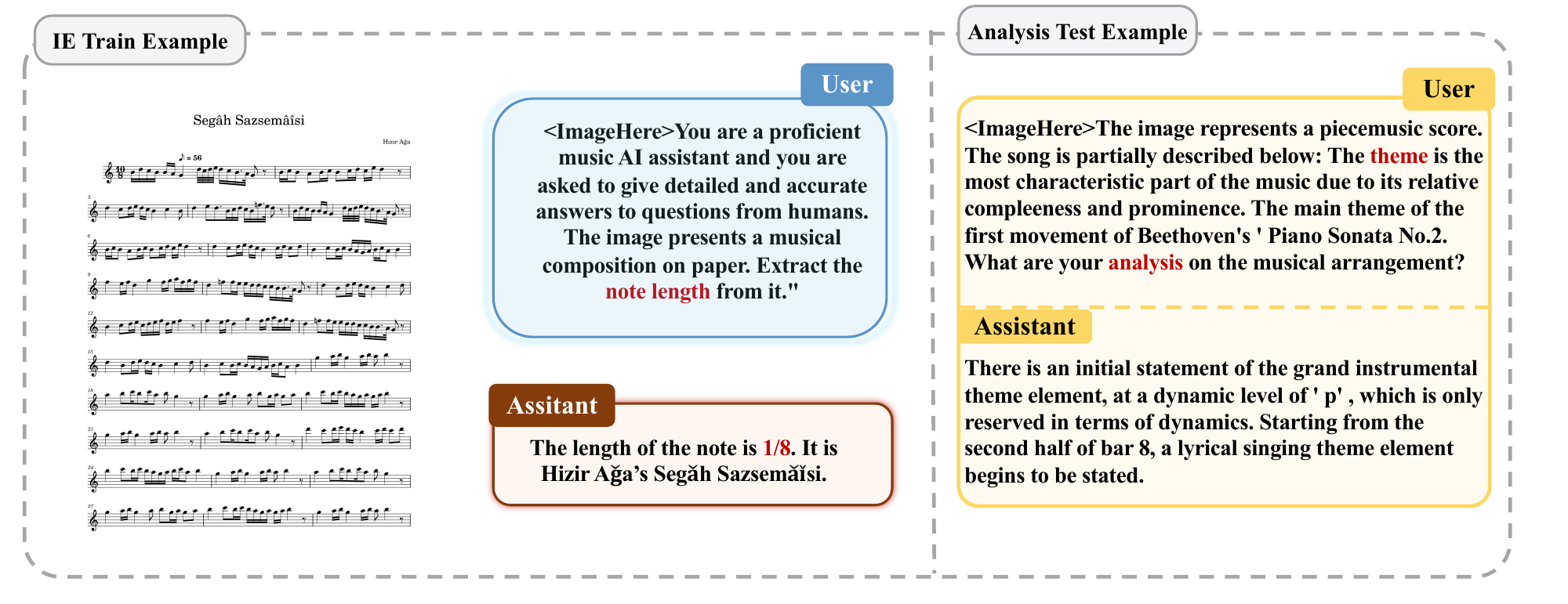}
    \caption{The left side of the figure shows an example of the information extraction task on the training dataset. The right side on the figure shows an example of music analysis for the test dataset.}
    \label{fig:data-example}
\end{figure*}

\section{NOTA Dataset}\label{section_dataset}
Our dataset is collected around three tasks: cross-modal alignment, music information extraction, and music notation analysis.
We choose to use ABC notation to represent music scores.
ABC notation encodes music into two parts: header and body. The first header is the reference number and the other headers are title T, time signature M, default note length L, key K, etc.  The body mainly includes notes, bar lines, and so on.
\paragraph{\textbf{Music Information Extraction}}\label{IE} In this task, we collect a total of 1,185,761 data entries. Music information extraction is divided into 6 subtasks: extracting ABC notation from corresponding images, and extracting specific information from the ABC notation, including T (tune title), K (key), L (unit note length), M (meter), and C (composer). We obtained 193,484 data entries from the ABC notation website, the vast majority of which are directly downloaded, and a small portion are scraped. After data cleaning, we only keep the ABC files that could generate the correct music score (we remove the original ABC file's comments, lyrics, and sequence numbers (X:)). We then transform ABC files into MusicXML files and use MuseScore4 to generate music score images from the MusicXML files.
Afterward, we divide each data entry into 6 data entries corresponding to 6 subtasks, resulting in 1,160,904 data entries. 

In order to test whether MLLMs have a special tendency towards certain regions, we additionally collect nearly 4000\textit{.krn} files from the internet, subsequently use the humdrum toolkit to convert them into ABC files, then filter and convert them into MusicXML files, generate music score from MusicXML files, and finally divide them into 6 extraction subtasks, obtaining a total of 24857 data entries with three regional labels: <China>, <Europe>, and <America>.

Each data sample includes the ABC notation information to extract, the corresponding music score images, the prompt used for extracting, and the gold answer.
Data examples are in Figure \ref{fig:data-example}.

\paragraph{Cross-modal Alignment}\label{align}
In this task, we obtain 29,116 data entries. We highlight portions of the music score images, expecting that MLLMs can understand and extract the corresponding ABC notation content. Each music score image has 2 to 4 highlighted sections. For a music score image $X_v$ and its associated content $X_c$, we sample a question $X_q$, which asks to extract the specific content of the image. With ($X_v$, $X_c$, $X_q$), we create a single-turn instruction-following example:
\begin{multline}
\text{Human : } <ImageHere> \; X_q \; X_v \; <STOP> \\
\text{Assistant : } X_c \; <STOP>
\end{multline}

\paragraph{Music Notation Analysis}\label{notation analysis}
This task includes analysis of score structure and musical styles.
In terms of score structure analysis, it involves systematic analysis of various musical elements such as structure, melody, harmony, tonality, rhythm, tempo, dynamics, texture, etc.
We integrate authoritative works on domestic and international music notation analysis. We obtain 250 questions on score structure analysis and 600 questions on musical style notation analysis. These questions cover the analysis of classic works from different countries (Germany, France, Italy, the UK, the United States, and so on) and different historical periods (from the Baroque period to the 20th century), involving various musical genres such as sonatas, symphonies, waltzes, and operas. Each data entry contains title, composer, the corresponding image, a description, and an analysis or structural breakdown.

Our dataset is divided into a train dataset and a test dataset. The train dataset has 998,976 samples, and the test dataset has 20,961 samples. More details are provided in Figure~\ref{dataset construction}.

\section{NotaGPT Training}\label{nota training}
We apply Mistral-7B~\cite{mistral} as the base large language model and CLIP~\cite{clip} as the vision encoder. Using the same network architecture as LLaVA~\citep{llava1,llava2}, the text model and the visual coder are connected through a linear projection layer.\label{c4} The model is first pre-trained with generalized domain multimodal datasets, which enables the model to understand images.
Our music understanding training is mainly in three stages: cross-modal alignment, music information extraction, and music notation analysis, as shown in Figure~\ref{fig:framework}.

\begin{table*}[h!]
\small
  \centering
  \small
  \begin{tabular}{c|ccccc|c}
    \toprule
    \textbf{Model} & \textbf{Author} & \textbf{Title} & \textbf{K} & \textbf{L} & \textbf{M} & \textbf{Avg}  \\
    \toprule
    $\text{CogAgent-Chat-hf~\citep{cogagent}}$ &15.98&75.43&9.94&2.36&21.11&24.97\\
    $\text{Cogvlm-Chat-hf~\citep{cogvlm}}$ &10.31&65.77&7.02&0.22&20.63&20.79\\
    $\text{VisualGLM-6B~\citep{visualGLM}}$ &0.05&5.32&32.78&0.00&29.27&11.24\\
    $\text{DeepSeek-VL-1.3B-Chat~\citep{deepseekvl}}$&15.98&0.11&4.75&0.00&22.84&8.74 \\
    $\text{DeepSeek-VL-7B-Chat~\citep{deepseekvl}}$&30.89&0.11&10.04&11.72&28.46&16.24 \\
    $\text{InstructBLIP-Vicuna-7B~\citep{instruct}}$ &0.43&5.67&7.67&0.00&1.84&3.12\\
    $\text{Yi-VL-6B~\citep{yi}}$ &46.27&17.82&10.37&9.13&5.02&17.72\\
    $\text{Yi-VL-34B~\citep{yi}}$ &60.85&0.22&13.55&14.36&11.18&20.03\\
    $\text{LLaVA-v1.5-7B~\citep{llava1}}$&54.81&25.16&11.56&11.50&28.54&26.31\\
    $\text{LLaVA-v1.5-13B~\citep{llava1}}$&6.86&34.23&4.7&0.59&28.22&14.92\\
    $\text{LLaVA-v1.6-Vicuna-7B~\citep{llava1}}$&38.88 &59.56&6.97&1.94&23.95&26.26\\
    $\text{LLaVA-v1.6-Vicuna-13B~\citep{llava1}}$&11.99&60.69&7.99&0.92&7.84&17.89\\
    $\text{LLaVA-v1.6-34B~\citep{llava1}}$&15.66&62.31&11.18&1.46&28.22&23.76\\
    $\text{MiniCPM-Llama3-V2\_5~\citep{minicpm}}$ &27.59&77.70 & 11.56 &9.72 &23.65 &25.04\\
    $\text{Qwen-VL~\citep{qwen}}$ &78.24&11.72&17.82&14.74&17.12&27.93\\
    $\text{Qwen-VL-Chat~\citep{qwen}}$ &72.08&0.38&13.44&14.36&16.25&23.30\\
    \toprule
    $\text{Gemini-pro-vision~\citep{gemini}}$ &51.83&69.03&15.08&13.02&21.87&33.34\\
    $\text{GPT-4V~\citep{gpt4}}$ &\textbf{82.24}&\textbf{77.95}&11.02&1.35&27.54&33.33\\
    $\text{NotaGPT-7B}$ &75.00&15.44&\textbf{80.45}&\textbf{85.26}&\textbf{83.08}&\textbf{67.84} \\
    \toprule
  \end{tabular}
  \caption{Evaluation results of music information extraction task from the training dataset. 'T' representing Title, 'K' for Key, 'L' for Unit Note Length, 'M' for Meter, and 'C' for Composer.\ref{IE}}
    \label{tab:main}
\end{table*}
\paragraph{Cross-modal Alignment}
At this stage, the primary goal is to achieve feature alignment between the musical notes depicted in music scores images and their textual representation in ABC notation.~Existing large vision models inherently lack this capability, as their pre-training does not include content specifically aligned with this requirement. Therefore, we have undertaken training modifications to enhance our model's performance. Specifically, we utilized the dataset introduced in section~\ref{align} to train the model. We have frozen the visual encoder and the language model components, focusing solely on training the two-layer MLP vision-language connector. This approach has enabled pre-alignment and endowed the model with the capability to recognize musical notes accurately.

\paragraph{Music Information Extraction}
Next, train the model to recognize the basic structure of music compositions and to extract relevant musical knowledge from images. Utilizing the training dataset described in section~\ref{IE}, we conducted fine-tuning of the entire model parameters while freezing the visual encoder component and training the remaining parts. Through this phase of training, the model's capability to extract musical information has significantly improved. It is now able to recognize fundamental elements of music scores such as beat types, note lengths, and key signatures from music score images.

\paragraph{Music Notation Analysis}
In the final phase, we fine-tuned the model using supervised fine-tuning, thereby enhancing its capability to understand and generate music. This phase involved using the 
 section~\ref{notation analysis} data to train the pre-trained projectors and the language model with full parameter adjustments. Post-training, the model has developed the ability to critically analyze music scores provided by users and perform complex tasks such as continuing a musical melody based on the preceding tune. 
\begin{figure}
    \centering
    \includegraphics[width=1\linewidth]{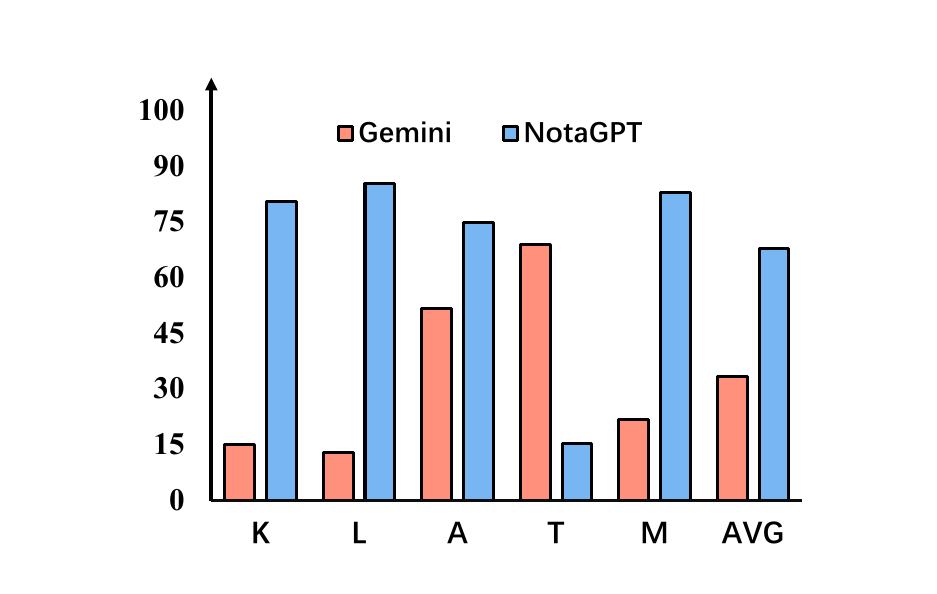}
    \caption{Extraction capabilities comparing between Gemini and NotaGPT-7B.}
    \label{Gemini}
\end{figure}
\begin{table}
\small
    \centering
\begin{tabular}{@{}lc@{}}
\toprule
Model & Levenshtein Distance  \\ \midrule
\multicolumn{2}{c}{\textsl{\# Generative MLLM}} \\ 
$\text{VisualGLM-6B}$ &643.72\\
\text{CogAgent-Chat}&730.65 \\
\text{DeepSeek-VL-1.3B-Chat}&316.85 \\
\text{DeepSeek-VL-7B-Chat}&308.27 \\
$\text{InstructBLIP-Vicuna-7B}$ &355.60\\
$\text{Yi-VL-6B}$ &561.47\\
$\text{Yi-VL-34B}$ & 522.07\\
$\text{LLaVA-v1.5-7B}$&667.08\\
$\text{LLaVA-v1.5-13B}$&147.47\\
$\text{LLaVA-v1.6-Vicuna-7B}$&807.75 \\
$\text{LLaVA-v1.6-Vicuna-13B}$&918.94 \\
$\text{LLaVA-v1.6-34B}$ &770.58\\
$\text{Qwen-VL}$ & 439.82\\
$\text{Qwen-VL-Chat}$ &625.16\\
\midrule
\multicolumn{2}{c}{\textit{\#Generative MLLM with api-token}} \\ 
$\text{Gemini-pro-vision}$ &354.30 \\
GPT-4V & 655.45  \\
\midrule
\multicolumn{2}{c}{\textit{\# Our Models}} \\
NotaGPT-7B&  \textbf{59.47}\\
\bottomrule
\end{tabular}
    \caption{Cross-modal alignment evaluation.}
    \label{edit-distance-result}
    \label{LevenshteinDistance}
\end{table}
\section{Experiments}
\subsection{Experiment Setup}
\paragraph{Baselines}~We comprehensively assess 17 MLLMs, including API-based models and open-source models. The API-based models contain GPT-4V~\citep{gpt4}, and Gemini ~\citep{gemini}. The open-source models contain LLaVA~\citep{llava1,llava2} series, VisualGLM~\citep{visualGLM}, Qwen-VL~\citep{qwen} series, and Yi-VL~\citep{yi} series.
\paragraph{Training Details}\label{c3}For pre-training, we utilized the alignment section~\ref{align} data conducting training 10 epoch with a learning rate of 2e-4. For supervised fine-tune training, we employed the train data in section~\ref{IE}, training 3 epochs with a learning rate of 2e-5 and a batch size of 32. All experiments are conducted on 8×80GB NVIDIA A100 SXM GPUs.\label{c5} 
\paragraph{Evaluation Details}~The temperature parameter was set to 0 to ensure deterministic output. For each model, we performed 3 separate evaluations using the GPT-4 API. The final score is determined by averaging the results from these 3 assessments.

\subsection{Evaluation Metrics}
\paragraph{Closed-set tasks.}
\textit{\textbf{(1)}}~For tasks such as \textit{music information extraction}, performance is assessed using the weighted extraction rate. They are questions with definitive answers such as music titles and note lengths. Given a response sequence \( R \) and an answer sequence \( A \) across a dataset of \( n \) queries, the overall success of the extractions can be defined as:
\vspace{-0.3cm}
\begin{equation}
\small
Extraction~~Rate= \sum_{i=1}^n \delta\left([A_i \subseteq R_i], 1\right)
\end{equation}
where \( \delta(x, y) \) is the Kronecker delta function, which equals 1 if \( x = y \) and 0 otherwise. The condition \( [A_i \subseteq R_i] \) evaluates to 1 if the answer sequence \( A_i \) is contained within the response sequence \( R_i \), and 0 otherwise. 

\textit{\textbf{(2)}}~Regarding the task of~\textit{converting images to ABC notation text}, we utilize the Levenshtein Distance~\cite{edit} as evaluation metric. It refers to the minimum number of single-character operations required to transform model responses into answer sequence. Let \(D\) be a matrix of size \((|R|+1) \times (|A|+1)\), where \(D[i][j]\) denotes the minimum edit distance between the first \(i\) characters of \(R\) and the first \(j\) characters of \(A\). The subsequent values of \(D\) are computed using the recurrence relation:
\begin{equation}
\small
D[i][j] = \min \left\{
\begin{array}{ll}
D[i-1][j] + 1 & (delet) \\
D[i][j-1] + 1 & (insert) \\
D[i-1][j-1] + cost & (substitute)
\end{array}
\right.
\end{equation}

where \(\text{cost}\) is \(0\) if the characters \(R[i-1]\) and \(A[j-1]\) are the same, and \(1\) otherwise.
\paragraph{Open-set tasks.} For \textit{notation analysis} tasks with open-ended answers, we used 2 type assessment:

\textit{\textbf{(1)}}Calculating using metrics. Our metrics are divided into two categories: semantic similarity and word matching. For semantic similarity, we use LSA, which measures the semantic similarity of text by computing the cosine similarity between vectors. For word matching, we use ROUGE-1, ROUGE-L, and METEOR, which respectively calculate the number of unigram matches, longest common subsequence matches, and synonym matches.


\textit{\textbf{(2)}}Scoring using LLM as an evaluator. As existing studies~\citep{gpt4-eval}~demonstrated, strong LLMs can be good evaluators. We compare the analysis generated by NotaGPT-7B with the analysis generated by other models, and have GPT-4 (text model) evaluate the analysis from both models. The evaluation considers both the music itself and the music's background. The evaluation of the music itself includes aspects such as musical language (melody, tonality, rhythm, musical terminology, etc.), technique application, and composition style. The evaluation of the music's background includes considerations of the social, historical, and cultural context, including the composer's milieu, the background of the composition, and the ideology of the creation. 
\begin{table*}[h!]
  \centering
  \small{
  \begin{tabular}{c|cccc|c}
    \toprule
    \textbf{Model} & \textbf{LSA} & \textbf{ROUGE-1} & \textbf{ROUGE-L} & \textbf{METEOR} & \textbf{Avg}
    \\
    \midrule
    $\text{InternVL-Chat-v1.5}$ & 14.96&19.71 & 13.32 &19.68 & 16.92\\
    $\text{InternVL-14B-224px}$ & 3.28&5.30 & 4.63 &4.18 &4.35\\
    $\text{VisualGLM-6B}$ &10.36&21.61&13.21&18.19 &15.84\\
    $\text{DeepSeek-VL-7B-base}$ & 9.92&16.43 &11.60 &13.81 &12.94\\
     $\text{InstructBLIP-Flan-T5-xl}$ & 9.38& 20.91 &15.28 & 14.57 &15.04\\
    $\text{InstructBLIP-Flan-T5-xxl}$ &7.64 &17.55 &12.32 &14.96 &13.12\\
    $\text{InstructBLIP-Vicuna-7B}$ & 8.28&22.23 &14.93 &16.74 &15.55\\
    $\text{InstructBLIP-Vicuna-13B}$ & 8.37&20.29 &14.18 &14.17&14.25\\
    $\text{MiniCPM-Llama3-V2\_5}$ &\textbf{16.26 }&20.72 & 13.36 &\textbf{20.83} &17.79\\
    $\text{Yi-VL-6B}$ &11.77&18.66&13.04&15.84 &14.83\\
    $\text{Yi-VL-34B}$ &12.47&19.44&13.20&17.18 &15.57\\
    $\text{Qwen-VL}$ &9.58&15.21 &10.37&12.56 &11.93\\
    $\text{Qwen-VL-Chat}$ & 9.66&16.80 &11.37 & 14.42 &13.06\\
    \midrule
    $\text{Gemini-pro-vision}$ &15.88&22.21&15.09&20.31&\textbf{18.37}\\
    $\text{GPT-4V}$ &14.03 & 18.49 &11.36 &19.94&15.96\\
    $\text{GPT4o}$ & 15.92& 18.27 &11.35 &20.26&16.45\\
    $\text{NotaGPT-7B}$ &12.46 &\textbf{22.63} &\textbf{15.53}&18.34 &17.24\\
    \toprule
  \end{tabular}
  }
  \caption{Comparisons of analysis and form Evaluation (\%). Part 1: Open-source models; Part 2: API-based models.}
  \label{music analysis}
\end{table*}
\section{Results}
Our experiment revolves around proving the effectiveness of NOTA in promoting music understanding. In order to enable the model to ultimately achieve music understanding, we have broken down the experiment into three sub-experiments: music information extraction, cross-modal alignment and music notation analysis.
Music information extraction only extracts the basic elements from the score image, such as author information, title, T, K, L, M and C.
Score image recognition builds upon the basic element extraction, further extracting the music score in ABC notation form.
Music analysis then, based on the extracted music score, conducts understanding and analysis, including score structure analysis and musical style analysis.

\begin{table*}[htbp]
\centering
  \small
\resizebox{0.75\textwidth}{!}{%
\begin{tabular}{c|c|ccc|ccc|c@{}}
\toprule
\multirow{2}{*}{\textbf{Model A}} & \multirow{2}{*}{\textbf{Type}} & \multicolumn{3}{c|}{\textbf{Musical styles}} & \multicolumn{3}{c|}{\textbf{Score Structures}} & \multirow{2}{*}{\textbf{C-Rate}} \\
 & & \textbf{A win} & \textbf{Tie} & \textbf{B win} & \textbf{A win} & \textbf{Tie} & \textbf{B win} & \\
\midrule
\multirow{2}{*}{InstructBLIP-Flan-T5-xxl} & w/ Info. & 5.00 & 33.50 & 61.50 & 1.34 & 33.56 & 65.10 & 96.56 \\
 & w/o Info. & 5.50 & 39.00 & 55.50 & 1.34 & 26.84 & 71.81 & 96.27 \\
\midrule
\multirow{2}{*}{InstructBLIP-Vicuna-7B} & w/ Info. & 1.00 & 25.00 & 74.00 & 2.68 & 32.89 & 64.43 & 98.28 \\
 & w/o Info. & 1.50 & 36.00 & 62.50 & 2.01 & 28.86 & 69.12 & 98.28 \\
\midrule
\multirow{2}{*}{InstructBLIP-Vicuna-13B} & w/ Info. & 1.00 & 26.50 & 72.50 & 1.34 & 30.87 & 67.79 & 98.85 \\
 & w/o Info. & 2.00 & 35.00 & 63.00 & 0.13 & 23.48 & 75.17 & 98.28 \\
\midrule
\multirow{2}{*}{InternVL-Chat-v1.5} & w/ Info. & 57.00 & 33.50 & 9.50 & 48.32 & 44.29 & 7.38 & 46.70 \\
 & w/o Info. & 35.00 & 49.00 & 16.00 & 26.84 & 55.70 & 17.44 & 68.48 \\
\midrule
\multirow{2}{*}{Qwen-VL} & w/ Info. & 24.50 & 45.00 & 30.50 & 16.11 & 39.60 & 44.30 & 79.08 \\
 & w/o Info. & 0.50 & 33.50 & 66.00 & 0.67 & 19.46 & 79.87 & 99.43 \\
\midrule
\multirow{2}{*}{VisualGLM-6B} & w/ Info. & 36.50 & 46.50 & 17.00 & 32.21 & 56.38 & 11.41 & 65.33 \\
 & w/o Info. & 14.00 & 46.50 & 39.50 & 11.40 & 40.93 & 47.65 & 34.67 \\
\midrule
\multirow{2}{*}{Yi-VL-6B} & w/ Info. & 36.00 & 40.50 & 23.50 & 30.20 & 49.66 & 20.14 & 66.47 \\
 & w/o Info. & 94.00 & 3.50 & 2.50 & 13.42 & 38.92 & 47.65 & 40.40 \\
\midrule
\multirow{2}{*}{GPT-4V} & w/ Info. & 69.50 & 25.00 & 5.50 & 55.70 & 33.56 & 10.74 & 36.39 \\
 & w/o Info. & 52.00 & 34.00 & 14.00 & 32.88 & 49.66 & 17.44 & 56.16 \\
\bottomrule
\end{tabular}%
}
\caption{Results of models generating music analysis, evaluated by GPT-4 (text model). \textit{Info.} means music background information, \textit{A win} means in GPT-4's view, model A's response is better than model B's as evaluated by GPT-4; \textit{tie} means the responses are equal; \textit{B win} means model B's response is better. \textit{C-Rate} means comparable rate between model B and model A.}
\label{beat}
\end{table*}


\subsection{Music Information Extraction Evaluation}
\paragraph{General comparison}
The evaluation results are presented in Table~\ref{tab:main}. We report the average extraction rate, with 23.53\% of the models showing an effective precision lower than 10\%. Additionally, 58.82\% of the models have an accuracy approximately between 10\% to 30\% , and only 17.64\% of the models achieve an accuracy exceeding 30\%. 
Overall,  NotaGPT-7B demonstrated the best performance among all the models evaluated, achieving an extracte rate of 67.84.
These findings highlight the challenges of the NOTA test dataset. 

 \paragraph{Comparative analysis} 
 Figure \ref{Gemini} illustrates the comparative performance of NotaGPT-7B and Gemini in several subcategories of an information extraction task. NotaGPT-7B significantly outperforms Gemini in the tasks of Author, K, L, and M, demonstrating the effectiveness of the training data. NotaGPT-7B does not perform very well on the title extraction task, and after analyzing it, we found that it is because it mistakenly extracts author information as title information.
 
 After training with the NOTA dataset, models of size 7B achieved substantial improvements in the categories K, L, and M, where performance was originally poor. These enhancements allowed them to surpass models of the same size and even those of larger sizes.

\subsection{Cross-modal Alignment Evaluation}
Table~\ref{edit-distance-result} presents the evaluation results. Overall, while high precision in music information extraction benefits cross-modal tasks, the relationship isn't simply linear. NotaGPT-7B consistently performs well, showcasing its strength in both extracting and aligning musical information. In contrast, while GPT-4V and Gemini-pro-vision score similarly in extraction tasks (around 33.34), they differ greatly in alignment accuracy, with Levenshtein distances of 655.45 and 354.30, respectively, suggesting that factors like model structure and optimization strategies also influence performance.
\subsection{Music Score Analysis Evaluation}
\paragraph{Metric evaluation}
Since the model's analysis and the standard answer cannot be completely identical, we evaluate the strength of the model's analysis capability of the recognized music score from semantic similarity and word matching.
    
\begin{figure}
    \centering
    \includegraphics[width=0.6\textwidth]{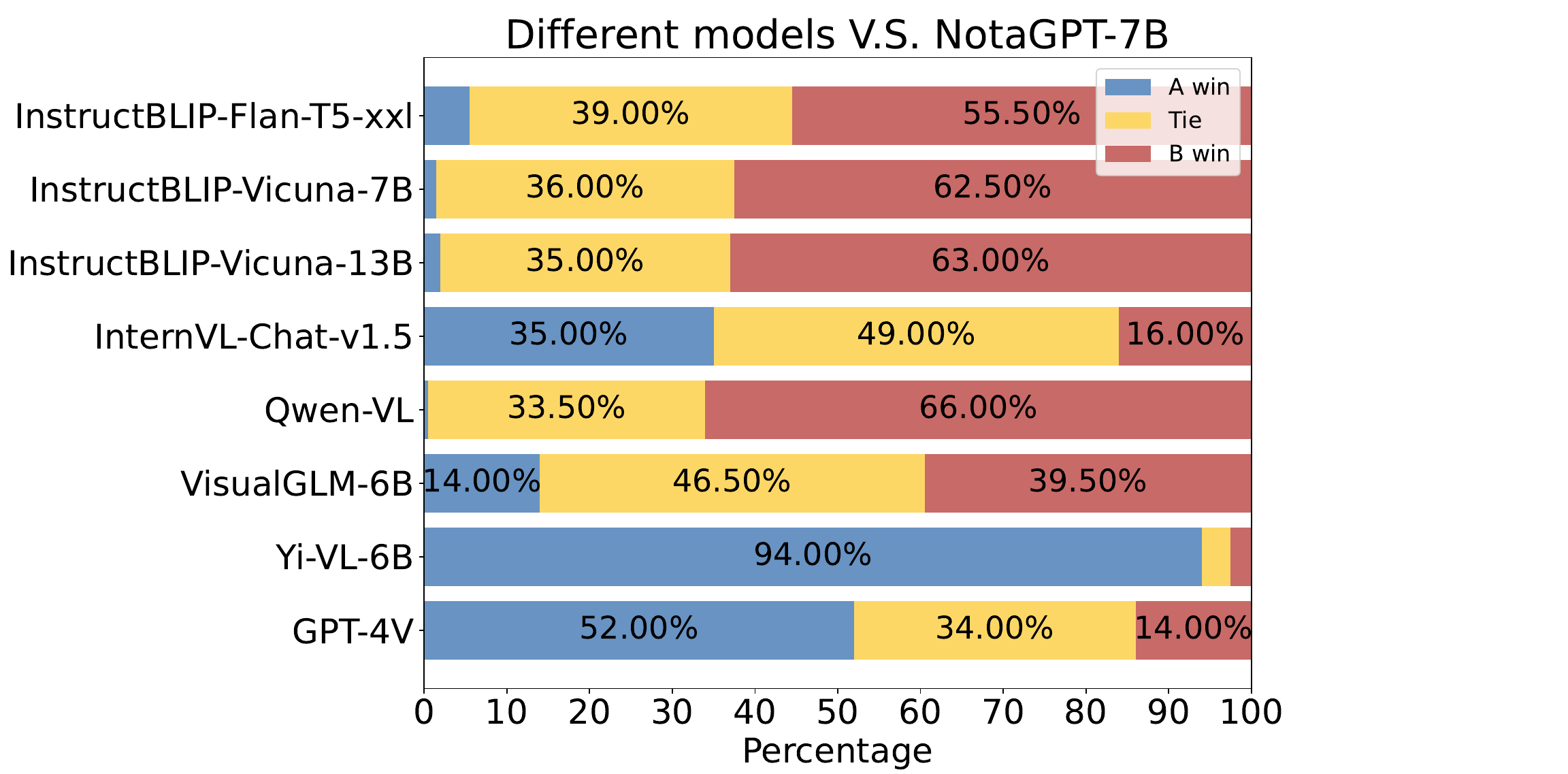}
    \caption{Visualization of evaluation results (w/o Info.) of all other models compared with our proposed NotaGPT model under GPT-4V.}
    \label{fig:analysis vs bar}
\end{figure}
From the results in Table~\ref{music analysis}, in terms of the LSA metric, the performance of NotaGPT-7B is stronger than most models, including some models with larger parameter sizes than 7B, only second to a few open-source models with larger parameter sizes, as well as API-based models. In word matching, NotaGPT-7B achieves SOTA performance on 2/3 of the metrics. 

NotaGPT-7B does not achieve the best performance on the LSA metric, on the one hand because the parameter size of NotaGPT-7B is only 7B, much smaller than the 25.5B of InternVL and the 34B of Yi-VL, which limits its capability; on the other hand, the base model of NotaGPT-7B does not use an instruction-tuned model like the Mistral-7B-Instruct series.
MiniCPM has similar size to NotaGPT-7B, based on the instruction-tuned model Llama3-8B-Instruct, whose capability is stronger than the base model Mistral-7B used by NotaGPT.

The results demonstrate the effectiveness of the NOTA test dataset, allowing the parameter-limited model NotaGPT-7B, after training, to outperform other models.

\paragraph{Analysis comparison}
Table~\ref{beat} contains the comparison between analysis of different models, and all the model B are NotaGPT-7B. Based on the results, NotaGPT-7B is better or on par with 75\% of the models. In comparison with most models, NotaGPT-7B's win rate is higher in the absence of music background information than with music background information. This performance can be attributed to NotaGPT-7B's training on a small set of music analysis data samples, which has endowed it with the capability to generally analyze musical scores and styles. It performs commendably even in prompts that lack background knowledge of the music piece.


\section{Conclusion}
In this study, we introduce NOTA, a large-scale music understanding dataset encompassing 3 tasks with over 1.1 million data entries. Based on the NOTA train dataset, we trained NotaGPT-7B, which demonstrates robust music notation understanding capability.
We further assess 17 multimodal models' capabilities in music understanding. The results show the constraints that are caused by the lack of multimodal music datasets, emphasizing the significance of the NOTA dataset. 

\section*{Limitations}
\label{limitation}
Although NOTA makes substantial advancement in developing effective music understanding datasets, we are aware of typical limitations in MLLMs, including hallucinations and shallow reasoning. Our future efforts will focus on improving the fidelity and dependability of these models.
\bibliography{custom.bib}
\bibliographystyle{acl_natbib}
\newpage
\appendix
\section{Appendix}

\subsection{Social Impact}\label{social impact}
The Nota-Eval dataset contains music from multiple regions and diverse cultural backgrounds. Not understanding the cultural context of the music may lead to misinterpretation of the music data, such as misreading the meaning and emotional expression of the music, as well as misjudging the characteristics and styles of the music.

\subsection{Region-Level Evaluation}

Table \ref{tab:region} presents the overall information extraction results for five information extraction tasks across 3 different regions using various models on our NOTA dataset. The experimental results indicate that the GPT-4V model significantly outperforms other models in music information extraction across different regions. For the five information extraction tasks in the regions of China and Europe, different models showed better performance compared to the America region. Additionally, there are noticeable differences in the information extraction capabilities of different models across the three regions. This suggests that different models have distinct preferences for understanding music from different regions, which may be related to the distribution of training data in these multimodal models.

\begin{figure}[b]
  \includegraphics[width=0.70\linewidth]{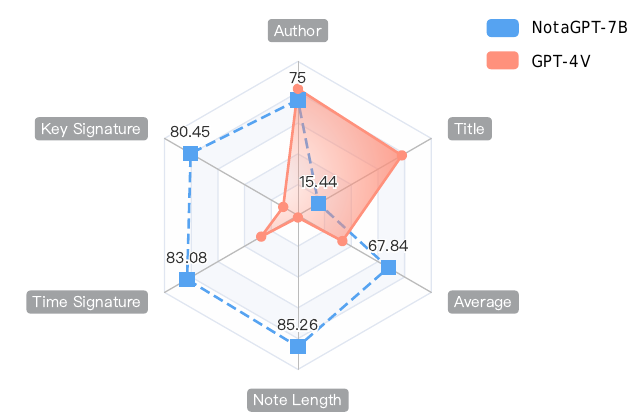}
  \centering
  \captionsetup{justification=centering}
  \caption{Comparing between GPT-4V and NotaGPT-7B.}

\end{figure}
\begin{table*}[h!]
  \centering
  \small
  \begin{tabular}{c|ccc|c}
    \toprule
    \textbf{Model} & \textbf{China} & \textbf{America} & \textbf{Europe} &  \textbf{Avg}  \\
    \toprule
    $\text{InternVL-14B-224px}$ &0.00&0.00&0.15&0.05 \\
    $\text{InternVL-Chat-V1.5}$ &0.48&5.56&1.81&2.61 \\
    $\text{VisualGLM-6B}$ &8.66&2.53&10.36&6.64\\
    $\text{DeepSeek-VL-1.3B-base}$ &7.51&0.64&8.09&5.03\\
    $\text{DeepSeek-VL-7B-base}$ &4.08&0.32&1.94&2.31\\
    $\text{InstructBLIP-Flan-T5-xl}$ &0.46&0.17&2.46&0.69\\
    $\text{InstructBLIP-Flan-T5-xxl}$ &1.03&0.00&5.24&1.36\\
    $\text{InstructBLIP-Vicuna-7B}$ &3.57&0.47&5.89&2.80\\
    $\text{InstructBLIP-Vicuna-13B}$ &1.08&0.12&2.65&0.98\\
    $\text{Yi-VL-6B}$ &0.14&0.03&0.19&0.11\\
    $\text{Yi-VL-34B}$&0.14&0.12&0.32&0.16\\
    $\text{MiniCPM-Llama3-V2\_5}$ &6.79&5.97&11.39&7.26\\
    $\text{Qwen-VL}$ &2.35&1.31&1.88&1.88\\
    $\text{Qwen-VL-Chat}$ &0.26&0.47&0.13&0.32\\
    \hline
    $\text{GPT-4V}$ &16.19&12.31&11.27&13.90\\
    \toprule
  \end{tabular}
    \caption{Region Bias Evaluation}
  \label{tab:region}
\end{table*}
\begin{figure*}[h]
    \centering
    \includegraphics[width=0.8\textwidth]{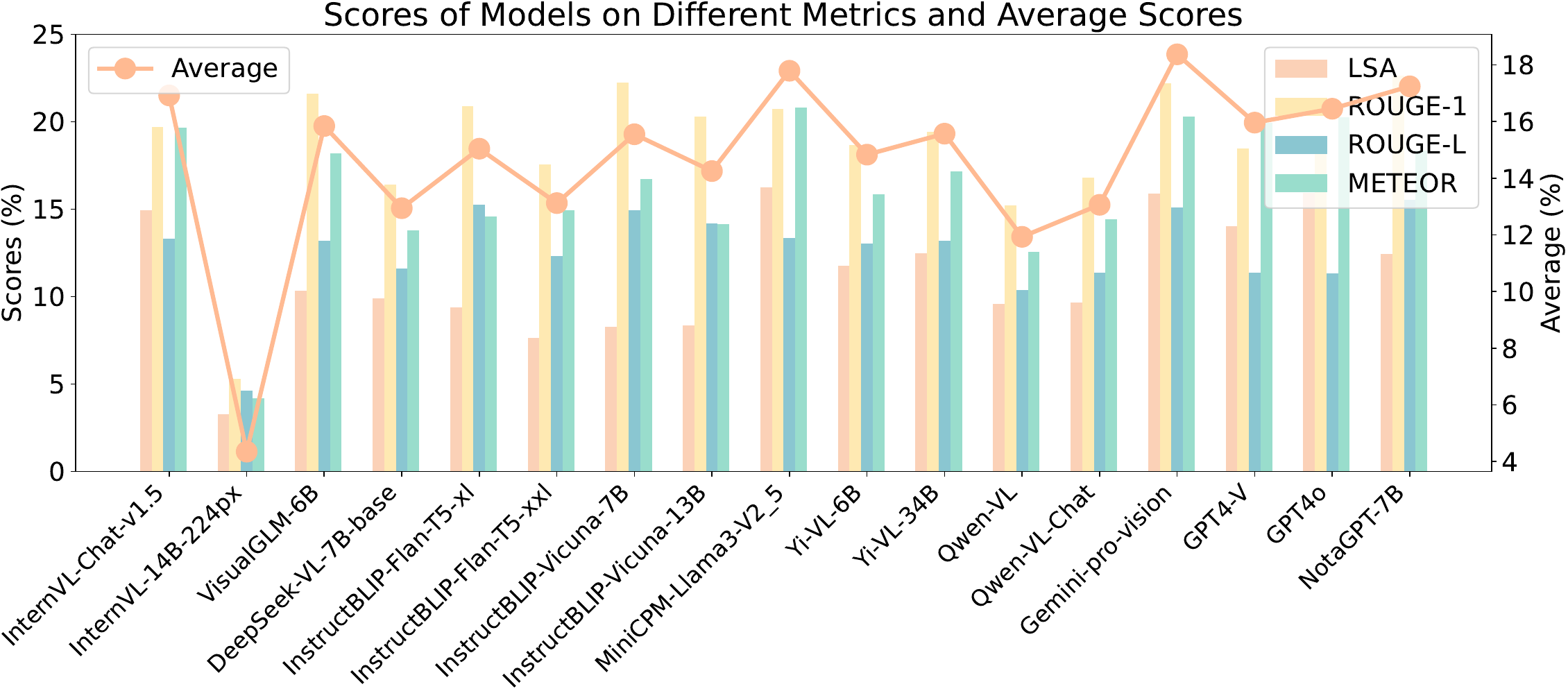}
    \caption{Music analysis figure.}
    \label{fig:example}
\end{figure*}

\subsection{Detailed Evaluation Metrics for Open-Set Tasks}
\textbf{Latent Semantic Analysis (LSA)} is a technique in natural language processing and information retrieval that analyzes relationships between a set of documents and the terms they contain by producing a set of concepts related to the documents and terms. LSA assumes that words that are close in meaning will appear in similar pieces of text. The core idea involves constructing a term-document matrix, which is then decomposed using singular value decomposition (SVD). The semantic similarity between texts is often measured using the cosine similarity between their vector representations. Let \( A \) be the term-document matrix, then LSA involves the following computation:

\[ A \approx U_k \Sigma_k V_k^T \]

where:
\begin{itemize}
  \item \( U_k \) represents the first \( k \) columns of \( U \),
  \item \( \Sigma_k \) is the top \( k \times k \) submatrix of \( \Sigma \),
  \item \( V_k^T \) is the first \( k \) rows of \( V^T \).
\end{itemize}

\textbf{ROUGE-1} is a metric used to evaluate automatic summarization and machine translation software, focusing specifically on the overlap of unigrams (single words) between the system-generated summary or translation and a set of reference summaries. The ROUGE-1 score is calculated by counting the number of unigrams in the generated text that match the unigrams in the reference text and then normalizing this number by the total number of unigrams in the reference text, providing a measure of recall. ROUGE-N is a metric for evaluating text summarization and machine translation quality by measuring the overlap of N-grams between system-generated summaries and reference summaries. Specifically, ROUGE-1 is a variant of ROUGE-N where N equals 1, meaning it calculates the overlap using unigrams (individual words). ROUGE-1 focuses on assessing the recall of single words, providing a basic measure of content overlap and is widely used due to its simplicity and effectiveness in capturing essential content accuracy. ROUGE-N can be represented as:

\[
\scalebox{0.95}{$
\text{Rouge-N} = \frac{\sum_{S \in \text{ReferenceSummaries}} \sum_{\text{gram}_n \in S} \text{Count}_{\text{match}}(\text{gram}_n)}{\sum_{S \in \text{ReferenceSummaries}} \sum_{\text{gram}_n \in S} \text{Count}(\text{gram}_n)}
$}
\]

\textbf{ROUGE-L} measures the longest common subsequence (LCS) between a system-generated summary or translation and a set of reference texts. It is particularly useful for evaluating the fluency and the order of the text in summaries and translations. The LCS does not require consecutive matches but is a sequence where each word is in the same order in both texts. The score is computed by dividing the length of the LCS by the total length of the reference sequence, providing insights into the overall text structure retention.It can be represented as:
\[
R_{\text{lcs}} = \frac{\text{LCS}(X, Y)}{m}
\]
\[
P_{\text{lcs}} = \frac{\text{LCS}(X, Y)}{n}
\]
\[
F_{\text{lcs}} = \frac{(1 + \beta^2) R_{\text{lcs}} P_{\text{lcs}}}{R_{\text{lcs}} + \beta^2 P_{\text{lcs}}}
\]

\textbf{METEOR}, or the Metric for Evaluation of Translation with Explicit ORdering, is a metric for evaluating machine translation output by aligning it to one or more reference translations. Unlike other metrics, METEOR accounts for exact word matches, synonymy, and stemming. It calculates scores based on the harmonic mean of precision and recall, weighted towards recall. The inclusion of synonyms and stemming allows METEOR to perform a more nuanced assessment of language use than simple exact matching. The METEOR score is calculated as follows:
\[ 
\text{METEOR} = F_{\text{mean}} \times (1 - \text{Penalty})
\]

where:
\[ 
F_{\text{mean}} = \frac{10 \cdot P \cdot R}{R + 9 \cdot P}
\]
\[ 
\hspace{1cm}
\scalebox{0.95}{$
\text{Penalty} = 0.5 \times \left( \frac{\text{number of chunks}}{\text{number of unigrams in candidate translation}} \right)^3
$}
\]

In these equations:
\begin{itemize}
  \item \( P \) is the precision,
  \item \( R \) is the recall,
  \item Chunks are contiguous sequences of words that are in the same order in both the candidate and the reference but are possibly interspersed with non-matching words.
\end{itemize}

\subsection{Author's statement and data license}
We undertake to assume all legal liability that may arise from the use of the dataset, in particular in relation to data infringement. This includes, but is not limited to, copyright infringement, privacy breaches or any other legal issues of any kind. With respect to the licensing of the data, we confirm that the dataset will be shared in compliance with applicable data protection regulations. The dataset will be licensed under a CC BY 4.0 license.

\subsection{The Role of Humans in Data Collection}
In the first two tasks, data was collected from electronic websites. The cleaning primarily involved dealing with some improperly formatted images and texts, as well as music pieces that had lost both author and title information, retaining only the melody.

In the final task, individuals were responsible for manually typing texts from over a dozen book publications. This included integrating authoritative works from both domestic and international sources in the fields of music appreciation, musical works analysis, and form analysis. Notable works included Yang Minwang's "New Compilation of World Famous Music Appreciation," Wu Zuqiang's "Form and Works Analysis," the "Norton Introduction to Music History" series, and Roger Kamien's "Music: An Appreciation," among dozens of seminal studies on Western musical works.
\end{document}